\documentclass[letterpaper, 10 pt, conference]{ieeeconf}

\IEEEoverridecommandlockouts                              % This command is only needed if 
\usepackage{theorem}
\usepackage{cite}
\usepackage{amsmath,amssymb,amsfonts}
\usepackage{graphicx}
\usepackage{xcolor}
\usepackage{comment}
\usepackage{bm}
\let\labelindent\relax
\usepackage{enumitem}
\usepackage{array}

\usepackage{algorithm}
\usepackage{algpseudocode}

\newtheorem{property}{Property}

\title{
Realtime Limb Trajectory Optimization for Humanoid Running Through Centroidal Angular Momentum Dynamics
}

\author{Sait Sovukluk, Robert Schuller, Johannes Englsberger, and Christian Ott% <-this % stops a space
\thanks{This project has received funding from the European Research Council (ERC) under the European Union’s Horizon 2020 research and innovation programme (grant agreement No. 819358).} %Use only for final RAL version
\thanks{Sait Sovukluk is with the Automation and Control Institute (ACIN), TU Wien, 1040 Vienna, Austria (e-mail: sovukluk@acin.tuwien.ac.at)}
\thanks{Robert Schuller and Johannes Englsberger is with the Institute of Robotics and Mechatronics, German Aerospace Center (DLR), 82234 Weßling, Germany (e-mail: Robert.Schuller@dlr.de, johannes.englsberger@dlr.de).}%
\thanks{Christian Ott is with the Automation and Control Institute (ACIN), TU Wien, 1040 Vienna, Austria, and also with the Institute of Robotics and Mechatronics, German Aerospace Center (DLR), 82234 Weßling, Germany (e-mail: christian.ott@tuwien.ac.at).}%
}

\begin{document}
\maketitle

\begin{abstract}
One of the essential aspects of humanoid robot running is determining the limb-swinging trajectories. During the flight phases, where the ground reaction forces are not available for regulation, the limb swinging trajectories are significant for the stability of the next stance phase. Due to the conservation of angular momentum, improper leg and arm swinging results in highly tilted and unsustainable body configurations at the next stance phase landing. In such cases, the robotic system fails to maintain locomotion independent of the stability of the center of mass trajectories. This problem is more apparent for fast and high flight time trajectories. This paper proposes a real-time nonlinear limb trajectory optimization problem for humanoid running. The optimization problem is tested on two different humanoid robot models, and the generated trajectories are verified using a running algorithm for both robots in a simulation environment.
\end{abstract}
\section{Introduction}
Humanoid robot locomotion is mainly composed of two parts. The first part and the main objective of locomotion is translating the overall body mass (center of mass) to a desired point and at a desired rate. The second part is preserving a proper posture that is suitable to the nature of the selected locomotion method. The posture objective also splits into two parts: keeping the torso upright and swinging the feet in the direction of locomotion to prepare for the next step. This paper aims to determine how to swing the limbs so that the robotic system is ready for the next step while preserving a proper posture.

As the influence of the contact forces on the centroidal dynamics is significant \cite{CMM1,holonomy}, motion generation for the center of mass (CoM) through a template or reduced models is very popular in the legged locomotion literature. These models are crafted to capture the ground reaction force patterns and centroidal dynamics well and are used to simplify some multi-body effects and limitations such that online control, planning, and decision-making are computationally feasible. The usage of centroidal or center of mass models in humanoid control appears in push recovery control through capture point \cite{CaptureP}; walking control through linear inverted pendulum (LIP) \cite{LIP}, divergent component of motion (DCM) \cite{DCM_walk}, and spring-loaded inverted pendulum (SLIP) based walking \cite{SLIP, SLIP_walk}; running control through SLIP \cite{wensingRun,sovuklukRun} and biologically inspired deadbeat control (BID) \cite{BID}. Control and planning through such models also appear in periodic jumping \cite{sovuklukJump}, bipedal backflip \cite{backflip} and are also widely used in quadruped locomotion literature in single rigid body dynamics (SRBD) form \cite{SRBD}. The locomotion control literature is covered extensively in \cite{patReview} and \cite{mpcRev}.

Even though the aforementioned methods showed success in simulation and hardware experiments, they do not capture the limb dynamics. All the methods assume the stance foot is placed at a specific location and do not include any information regarding the torso orientation, swing leg trajectories, and arm trajectories. Whole-body controllers handle this part of the locomotion control and require the user to select and tune these uncaptured trajectories. In the case of the slow pace and small step size walking, which covers the walking of almost all humanoid robots in the current market, the effect of limb swing is not that apparent mainly for two reasons. First, by the nature of walking behaviors, at least one foot is always in contact with the ground. Hence, the ground reaction forces can continuously be used for regulation as long as the friction cone allows. Second, the swing leg velocity is small due to the high step times and small foot displacements. Hence, the effect of leg swinging on the overall posture and dynamics is small. Such behaviors may not even require an arm swing behavior to regulate the angular momentum. As the walking gets more dynamic and the step length increases, the swing leg moves faster and affects on the overall dynamics more significantly. Such behaviors require more precise trajectory tuning both for the swing leg and arms. This problem is addressed in \cite{robert} through an optimization for the overall centroidal angular momentum target value.

In the case of running, which is the target of this paper, the aforementioned conditions get even more challenging. As the flight phase does not involve any contact with the ground, limb movement during this phase drastically affects the overall posture due to the conservation of angular momentum. Such behaviors require either very well-tuned limb swing trajectories or have to be limited with very short-lasting flight phases. This problem is more apparent while running through random environments where, due to lack of periodicity, each step may require different running trajectories along with their well-tuned limb swing characteristics. Unprecise tuning of limb trajectories results in ill-defined postural configurations and high body rotational velocities at the touchdown, which cannot be dissipated by the time and friction cone limited stance phase control. Independent from the desired center of mass trajectory tracking performance, such systems inevitably fail to sustain their motions either immediately or as a result of the accumulated body orientation error. We demonstrate such behaviors in the supplemental video.

This paper focuses on online nonlinear constrained limb trajectory optimization through centroidal angular momentum dynamics for humanoid running. The objective of the optimization is to determine the liftoff configuration and flight phase joint trajectories such that at the next touchdown, the body orientation is minimal. The proposed optimization structure is independent of centroidal running trajectory generation methods. It requires only the desired stance leg position w.r.t. the CoM location at the moment of liftoff and touchdown. While doing so, we exploit some properties of the centroidal momentum matrix (CMM), reduce the size of the problem by some minimal simplifications, and construct a nonlinearly constrained optimization problem that can be solved in real-time. We implement our nonlinear optimization solver for the best efficiency and customization. Lastly, we verify our method in a simulation environment on two different robots. We also implement SLIP-based humanoid running trajectories and verify that the optimized flight phase joint trajectories result in minimal body orientation in the next touchdown phase. The running trajectories are constructed for long flight phases, even longer than the stance phase, such that they are more challenging and harder to control. Lastly, we share the computation time report and show that online implementation is possible.
\section{System Dynamics}
Let $\bm{q}$ be a set of configuration variables and $\bm{\nu} = (\bm{\nu}_{b}, \bm{\nu}_{j})$ be the generalized velocity where $\bm{\nu}_{b} = (\bm{v}_{b},\bm{\omega}_{b}) \in \mathbb{R}^{6}$ is the linear and angular velocity of the floating base and $\bm{\nu}_{j} \in \mathbb{R}^{n}$ is the generalized velocity of the joints. The well-known robotic system dynamics results in
\begin{equation} \label{systemDyn}
\underbrace{
\begin{bmatrix}
    \bm{M}_{bb} & \bm{M}_{bj}\\
    \bm{M}_{jb} & \bm{M}_{jj}
\end{bmatrix}}_{\bm{M}}
\begin{bmatrix}
    \dot{\bm{\nu}}_{b}\\
    \dot{\bm{\nu}}_{j}
\end{bmatrix}
+ \bm{C}(\bm{q},\bm{\nu})\bm{\nu} + \bm{\tau}_{g}(\bm{q}) = 
\begin{bmatrix}
    \bm{0}\\
    \bm{\tau}
\end{bmatrix}
+
\bm{J}_{c}(\bm{q})^{\top} \bm{f}_{c},
\end{equation}
where $\bm{f}_{c}\in\mathbb{R}^{3n_{c}}$ is the vector of contact forces for $n_{c}$ number of contacts. The first six rows of \eqref{systemDyn} correspond to the floating base dynamics and are the underactuated portion of the system dynamics, i.e., the base dynamics cannot directly be driven by the instantaneous joint torques. Hence, the base movement is determined by the contact forces and inertial couplings.

Define $\bm{h}_{G} = [\bm{l}_{G}; \bm{k}_{G}] \in \mathbb{R}^{6}$ as the centroidal momentum, which is a combination of translational $\bm{l}_{G}\in\mathbb{R}^{3}$ and rotational (angular) $\bm{k}_{G}\in\mathbb{R}^{3}$ momentum. Then, based on \cite{holonomy} and \cite{CMM2}, the first six rows of \eqref{systemDyn} are equivalent to
\begin{equation} \label{CMrate}
    \dot{\bm{h}}_{G} = 
    \begin{bmatrix}
        \dot{\bm{l}}_{G} \\
        \dot{\bm{k}}_{G}
    \end{bmatrix}
    = 
    \begin{bmatrix}
        m\bm{g} \\
        \bm{0}
    \end{bmatrix}
    +
    \sum_{i=1}^{n_{c}}
    \begin{bmatrix}
        \bm{f}_{c,i} \\
        (\bm{p}_{c,i} - \bm{p}_{\text{CoM}}) \times \bm{f}_{c,i}
    \end{bmatrix},
\end{equation}
where $\bm{g}\in\mathbb{R}^{3}$ is the gravitational acceleration vector, $\bm{f}_{c,i}\in\mathbb{R}^{3}$ is the contact force vector originating from $\bm{p}_{c,i}\in\mathbb{R}^{3}$, $\bm{p}_{\text{CoM}}\in\mathbb{R}^{3}$ is the position of center-of-mass, and $m$ is the total mass. The centroidal momentum dynamics \eqref{CMrate} shows the significance of the contact forces on the body dynamics and is one of the main motivations of template model based locomotion approaches \cite{patReview,CMM1}. The centroidal momentum is related to the generalized velocities through $\bm{h}_{G} = \bm{A}_{G}(\bm{q}) \bm{\nu}$ where $\bm{A}_{G} \in \mathbb{R}^{6 \times (n+6)}$ is the centroidal momentum matrix (CMM) \cite{CMM1,CMM2}.
\section{Properties of the Centroidal Angular Momentum Matrix}
In order to focus our discussion on the angular portion of the CMM, we decompose the linear and angular portions:
\begin{equation}
    \bm{A}_{G} = 
    \begin{bmatrix}
        \bm{A}_{l} \in \mathbb{R}^{3 \times (n+6)} \\
        \bm{A}_{k} \in \mathbb{R}^{3 \times (n+6)}
    \end{bmatrix}.
\end{equation}
The centroidal angular momentum matrix $\bm{A}_{k}$ can also be decomposed into
\begin{equation} \label{Ak}
    \bm{A}_{k} = 
    \begin{bmatrix}
        \bm{A}_{v}\in\mathbb{R}^{3 \times 3} & \bm{A}_{\omega}\in\mathbb{R}^{3 \times 3} & \bm{A}_{j}\in\mathbb{R}^{3 \times n}
    \end{bmatrix},
\end{equation}
where $\bm{A}_{v}$, $\bm{A}_{\omega}$, and $\bm{A}_{j}$ represent body translational, body rotational, and joint velocity portions, respectively.
\begin{property}
$\bm{A}_{v}$ is a zero matrix.
\end{property}
\begin{proof}
Let $^{G}\bm{R}_{B} \in SO(3)$ be a rotation matrix from the body frame to the centroidal (CoM) frame. Additionally, let $^{B}\bm{X}_{G}^{T} \in \mathbb{R}^{6 \times 6}$ be momentum transformation from the floating base body frame to the centroidal frame \cite{handbookDynamics}:
\begin{equation}
    ^{B}\bm{X}_{G}^{T} = 
    \begin{bmatrix}
        ^{G}\bm{R}_{B} & \bm{0}_{3 \times 3}\\
        ^{G}\bm{R}_{B} \bm{S}(^{B}\bm{p}_{G})^{\top} & ^{G}\bm{R}_{B}
    \end{bmatrix}.
\end{equation}
The centroidal momentum matrix can be calculated as \cite{CMM2}
\begin{equation}
    \bm{A}_{G} = 
    \begin{bmatrix}
        \bm{A}_{l} \in \mathbb{R}^{3 \times (n+6)} \\
        \bm{A}_{k} \in \mathbb{R}^{3 \times (n+6)}
    \end{bmatrix}
    = {^{B}\bm{X}_{G}^{T}}
    \begin{bmatrix}
        \bm{M}_{11} & \bm{M}_{12} & \bm{M}_{13} \\
        \bm{M}_{21} & \bm{M}_{22} & \bm{M}_{23}
    \end{bmatrix},
\end{equation}
where the rightmost matrix is the first six rows of the mass matrix: $\{ \bm{M}_{11},\bm{M}_{12},\bm{M}_{21},\bm{M}_{22} \} \in \mathbb{R}^{3 \times 3}$ and $\{ \bm{M}_{13},\bm{M}_{23} \} \in \mathbb{R}^{3 \times n}$. Similarly, decomposing angular portion of the centroidal momentum matrix as 
\begin{equation*}
    \bm{A}_{k} = 
    \begin{bmatrix}
        \bm{A}_{v}\in\mathbb{R}^{3 \times 3} & \bm{A}_{\omega}\in\mathbb{R}^{3 \times 3} & \bm{A}_{j}\in\mathbb{R}^{3 \times n}
    \end{bmatrix}
\end{equation*}
results in
\begin{equation} \label{ApA1}
    \bm{A}_{v} = {^{G}\bm{R}_{B}} \bm{S}(^{B}\bm{p}_{G})^{\top} \bm{M}_{11} + {^{G}\bm{R}_{B}} \bm{M}_{21}.
\end{equation}
Similarly, based on \cite{handbookDynamics} and \cite{CMM2}
\begin{equation} \label{ApA2}
    \begin{bmatrix}
        \bm{M}_{11} & \bm{M}_{12} \\
        \bm{M}_{21} & \bm{M}_{22}
    \end{bmatrix}
    =
    \begin{bmatrix}
        m_{t} \bm{1}_{3 \times 3} & m_{t}\bm{S}({^{B}\bm{p}_{G}})^{\top} \\
        m_{t}\bm{S}({^{B}\bm{p}_{G}}) & \bar{\bm{I}}_{B}
    \end{bmatrix},
\end{equation}
where $\bar{\bm{I}}_{B}(\bm{q})\in\mathbb{R}^{3 \times 3}$ is the overall Cartesian inertia about the floating base body frame. Finally, the substitution of $\bm{M}_{11}$ and $\bm{M}_{21}$ from \eqref{ApA2} into \eqref{ApA1} yields
\begin{equation} \label{proof0}
    \bm{A}_{v} = m_{t}{^{G}\bm{R}_{B}} \bm{S}(^{B}\bm{p}_{G})^{\top} + m_{t}{^{G}\bm{R}_{B}} \bm{S}({^{B}\bm{p}_{G}}) = \bm{0}_{3 \times 3}.
\end{equation}
\end{proof}
The important implication of Property~1 is that the centroidal angular momentum dynamics is completely decoupled from the translational body velocities. Knowing the initial condition, joint trajectories, and flight time, the body orientation evolution is the same for any set of translational body velocities. Hence, the resultant optimization problem does not require any translational trajectory information and is completely independent of running models.

\begin{property}
$\bm{A}_{\omega}$ is always invertible.
\end{property}
\begin{proof}
Invertibility of $\bm{A}_{\omega}$ is intuitively apparent and has a similar proof. Similar to \eqref{ApA1}
\begin{equation} \label{ApB1}
    \bm{A}_{\omega} = {^{G}\bm{R}_{B}} \bm{S}(^{B}\bm{p}_{G})^{\top} \bm{M}_{12} + {^{G}\bm{R}_{B}} \bm{M}_{22}.
\end{equation}
Substitution of $\bm{M}_{12}$ and $\bm{M}_{22}$ from \eqref{ApA2} into \eqref{ApB1} yields
\begin{equation}
    \bm{A}_{\omega} = m_{t}{^{G}\bm{R}_{B}} \bm{S}(^{B}\bm{p}_{G})^{\top} \bm{S}({^{B}\bm{p}_{G}})^{\top} + {^{G}\bm{R}_{B}} \bar{\bm{I}}_{B}.
\end{equation}
where $\bar{\bm{I}}_{B} = \bar{\bm{I}}_{\text{com}} + m_{t}\bm{S}(^{B}\bm{p}_{G})\bm{S}(^{B}\bm{p}_{G})^{\top}$ \cite{handbookDynamics} and $\bar{\bm{I}}_{com}$ represents the overall inertia with respect to the center of mass. Consequently
\begin{equation} \label{ApB2}
    \bm{A}_{\omega} = {^{G}\bm{R}_{B}} \bar{\bm{I}}_{\text{com}}.
\end{equation}
Hence, $\bm{A}_{\omega}$ is always invertible as it is a product of two invertible matrices.
\end{proof}
The invertibility of $\bm{A}_{\omega}$ implies that for a given centroidal angular momentum and joint velocities, there exists a finite body rotational velocity at any system configuration. As this work focuses on estimating the body's rotational velocity and minimizing its integration during the flight phase, Property~2 ensures that there exists a solution to the calculation for any joint configuration and velocity.
\section{Flight Phase Body Orientation Dynamics}
The first obvious observation for the flight phase is the conservation of angular momentum. As the contact force $\bm{f}_{c}$ is a zero vector, there is no external force other than gravity acting on the system, i.e., $\dot{\bm{k}}_{G} = \bm{0}_{3 \times 1}$. Hence, during the flight phase, the body orientation cannot directly be controlled by the actuators. It can only be controlled by the coupling effects of the actuated links and this information is embedded inside the $\bm{A}_{j}$ matrix. Define $\bm{k}_{Gf} \in \mathbb{R}^{3}$ to be the centroidal angular momentum of the system during the flight phase such that:
\begin{equation*}
    \bm{k}_{Gf} = \begin{bmatrix} \bm{A}_{v} & \bm{A}_{\omega} & \bm{A}_{j} \end{bmatrix}
    \begin{bmatrix} \bm{v}_{b} \\ \bm{\omega}_{b} \\ \bm{\nu}_{j} \end{bmatrix}.
\end{equation*}
Due to the conservation of angular momentum, $\bm{k}_{Gf}$ is constant throughout the flight phase. The consequent body orientation dynamics yields:
\begin{equation*}
    \bm{\omega}_{b} = \bm{A}_{\omega}^{-1}(\bm{k}_{Gf}-\bm{A}_{v}\bm{v}_{b}-\bm{A}_{j}\bm{\nu}_{j}),
\end{equation*}
where $\bm{A}_{\omega}$ is always invertible by Property~2.
Furthermore, due to Property~1, the body orientation dynamics results in
\begin{equation} \label{oDyn}
    \bm{\omega}_{b} = \bm{A}_{\omega}^{-1}(\bm{k}_{Gf}-\bm{A}_{j}\bm{\nu}_{j}).
\end{equation}
The body orientation dynamics \eqref{oDyn} implies a few important aspects. First and most apparent, it shows that the body orientation can be implicitly controlled through the actuated link joints. Second, during the flight phase, the orientation dynamics is completely decoupled from the translational trajectory. This decoupling allows the body rotational velocity estimation to be independent of the robotic system's flight phase trajectory.
\section{Optimization Problem Formulation}
\begin{figure}[t!]
\centerline{\includegraphics[width=\columnwidth]{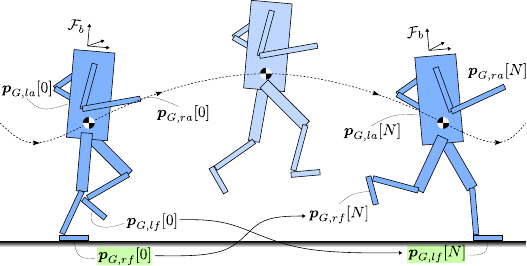}}
\caption{Illustration of humanoid running from liftoff to the consecutive touchdown. The green variables (liftoff and touchdown point w.r.t. CoM) are given by any running or jumping algorithm and assumed to be known. Uncolored parameters are not captured by the reduced models and require manual selections. The objective of optimization is to eliminate these selections.}
\label{confFig}
\end{figure}

As discussed in the introduction, centroidal model-based running and jumping planners fall short of identifying how the body, arm, and swing leg trajectories should evolve. They usually capture only the essence of the locomotion, i.e., center of mass, stance foot, and contact force evolution of the system. However, as shown in Fig.~\ref{confFig}, humanoid robots require much more to control. The stance foot location just before the liftoff phase is known. On the other hand, since the stance leg becomes the next step's swing leg, the user has to define an appropriate trajectory for its evolution. The same is valid for the swing leg. At the liftoff moment, the swing leg configuration is unknown and must be decided manually. However, at the next touchdown moment, the same leg becomes the stance leg and should be placed at a known location with respect to the center of mass. Nothing is given for arm evolution. They should be used to regulate the robot's posture. This section works on these unknown aspects and formulates an optimization problem to decide the limb evolution of the system.

Define $\bm{q}^{d}:\mathbb{R}\rightarrow\mathbb{R}^{n}$ to be a vector of polynomials for the desired actuated joint evolution of the system for the flight phase. Hence, $\bm{q}^{d}(0)$ and $\bm{q}^{d}(t_{f})$ represent the desired joint position at the beginning and end of the flight phase, respectively. Similarly, define $\bm{\theta}_{b}$ to be the body orientation. Hence, for a given initial body orientation $\bm{\theta}_{b}(0)$ the consequent final touchdown configuration integrated:
\begin{equation} \label{backInteg}
    \bm{\theta}_{b}(t_{f}) = \bm{\theta}_{b}(0) + \int_{0}^{t_f}{T(\bm{\theta}_{b}(t),\bm{\omega}_{b}(t))}dt,
\end{equation}
where function $T$ is a mapping from the angular velocities to the body configuration rate (for example Euler or Quaternion rate).

An algorithm that performs a discrete summation of the nonlinear integrator \eqref{backInteg} through $N$ number of sampling points is shown in Alg.~\ref{alg:integration}. Starting from a given (desired) liftoff base orientation and rotational velocity, the algorithm integrates for the touchdown configuration. The ``computeCentroidalMap'' function in the algorithm calculates the forward kinematics, center of mass location, and centroidal momentum matrix together (see documentation of Pinnochio \cite{pinocchioweb} for ``computeCentroidalMap''). The given algorithm will construct a nonlinear cost function for the optimization.
\begin{algorithm} [t!]
\caption{Integration of base orientation}\label{alg:integration}
\begin{algorithmic} [1]
\State $\bm{\theta} \gets \bm{\theta}(0)$\;
\State $t \gets 0$\;
\State $\bm{\omega}_{b} \gets \bm{\omega}_{b}(0)$\;
\State $(\bm{A}_{k}, \bm{p}_{G}) = computeCentroidalMap(\bm{\theta}_{0},\bm{q}^{d}(0))$\;
\State  $\bm{k}_{Gf} = \bm{A}_{k}\bm{\nu}$\;
\While{$t < t_{f}$}
\State  $\bm{\theta} \gets \bm{\theta} + T(\bm{\omega}_{b}(t))*t_{f}/N$\;
\State  $t \gets t + t_{f}/N$\;
\State  $(\bm{A}_{k}, \bm{p}_{G}) = computeCentroidalMap(\bm{\theta},\bm{q}^{d}(t))$\;
\State  $\bm{\omega}_{b} = \bm{A}_{\omega}^{-1}(\bm{k}_{Gf}-\bm{A}_{j}\dot{\bm{q}}^{d}(t))$\;
\EndWhile
\State  $\bm{\omega}_{b}(t_{f}) = \bm{\omega}_{b}$\
\end{algorithmic}
\end{algorithm}

\begin{figure}[t!]
\centerline{\includegraphics[width=\columnwidth]{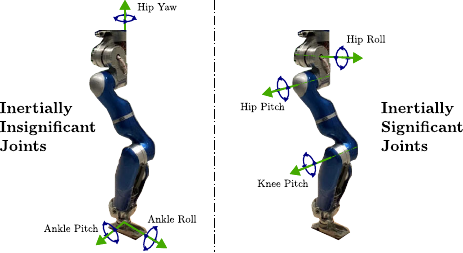}}
\caption{Leg joints of a typical humanoid robot. The left figure shows the joints that do not have significant effects on the inertial shape or angular momentum of the robot. On the other hand, the right figure shows the joints that both translate and rotate significant potions of inertia. As an example, the image shows TORO's right leg \cite{toro} and the joint locations are placed to illustrate the most common and intuitive configurations.}
\label{jointFig}
\end{figure}

The computation cost of the optimization can be reduced via some reasonable assumptions. Typical leg joints of a humanoid robot are shown in Fig.~\ref{jointFig}. The ankle link is a relatively small component of the robot and constitutes an insignificant portion of the total inertia. Since its body is a small mass with a very short link length, rotation of ankle joints has a negligible effect on the overall inertia shape and angular momentum. Hence, during the optimization, these joints can be assumed not to move, i.e., remain at a constant angle. This assumption still accounts for ankle inertia and mass but neglects the inertia change with respect to the ankle joint angle. Similarly, the hip yaw joint is also another joint that can be simplified. Even though this joint drives the whole leg, as the mass is distributed around its rotation axis, the inertia that this joint drives is still comparably small. Another important aspect of this joint is that, as seen in Fig.~\ref{confFig}, this joint is not very active during running and jumping but only during a change of direction. Hence, assuming this joint will stay at its default position during the flight phase is also a reasonable assumption. During the direction change step, where the support leg rotates to the new heading angle during the flight phase, its rotational effect can be accounted for in the optimization. Similar justifications can also be made for the arm joints, e.g., wrist and shoulder yaw joints. On the other hand, hip roll, hip pitch, shoulder pitch, and knee joints are highly inertial and have an important effect on the overall inertial shape. Additionally, these joints move in a wide range at high velocities and generate a significant portion of angular momentum. Hence, these joints are the main focus of the optimization

Let $\bm{\Gamma} \in \mathbb{R}^{n \times (m+1)}$ be the collection of degree $m$ polynomial constants for all desired joint trajectories such that
\begin{equation}
    \bm{q}^{d} = \underbrace{\begin{bmatrix}
        ^{1}\gamma_{m} & ^{1}\gamma_{m-1} & \cdots & ^{1}\gamma_{1} & ^{1}\gamma_{0}\\
        \vdots & \vdots & \vdots & \vdots & \vdots\\
        ^{n}\gamma_{m} & ^{n}\gamma_{m-1} & \cdots & ^{n}\gamma_{1} & ^{n}\gamma_{0}\\
    \end{bmatrix}}_{\bm{\Gamma}}
    \underbrace{\begin{bmatrix} t^{m} \\ t^{m-1} \\ \vdots \\ t \\ 1 \end{bmatrix}}_{\bm{t}}.
\end{equation}
From this point, we label the upcoming touchdown leg (left foot in Fig.~\ref{confFig}) and the upcoming swing leg (right foot in Fig.~\ref{confFig}) as stance and swing foot, respectively.
\subsection{Cost Function}
The objective of the cost function is to minimize the body orientation at the next touchdown moment:
\begin{equation} \label{costFunc}
    \min_{\bm{\Gamma}}{f_{\text{cost}} =||\bm{\theta}_{b}(t_{f})||}.
\end{equation}
From a given liftoff body orientation, the optimizer modifies the joint trajectories through the polynomial gains such that there is minimal body orientation at the next touchdown. The base orientation throughout the flight phase in \eqref{costFunc} can be calculated through Algorithm~\ref{alg:integration}.
\subsection{Constraints}
\begin{enumerate}
    \item Setting ankle, wrist, hip yaw, and arm yaw trajectories to zero polynomials:
        \begin{equation}
            \begin{bmatrix}
                ^{\text{ankle}} \bm{\Gamma} \\ ^{\text{wrist}} \bm{\Gamma} \\ ^{\text{hipYaw}} \bm{\Gamma} \\ ^{\text{armYaw}} \bm{\Gamma}
            \end{bmatrix}
            =
            \begin{bmatrix}
                \bm{0}_{4 \times 4} \\ \bm{0}_{4 \times 4} \\ \bm{0}_{2 \times 4} \\ \bm{0}_{2 \times 4}
            \end{bmatrix}.
        \end{equation}
    In case of a direction change, during the transient step, the hip yaw joint trajectory of the related leg can be encoded here to take account of the inertial shape changes.
    \item Enforcing the touchdown leg position for the upcoming stance phase through a forward kinematic constraint:
        \begin{equation}
             \bm{p}(^{\text{stance}}\bm{\Gamma}\bm{t}(t_{f}))-\bm{p}_{G}(t_{f}) = \bm{p}_{G,\text{stance}}^{d}.
        \end{equation}
    \item Enforcing the liftoff swing leg position through a forward kinematic constraint:
        \begin{equation}
             \bm{p}(^{\text{swing}}\bm{\Gamma}\bm{t}(0))-\bm{p}_{G}(t_{f})= \bm{p}_{G,\text{swing}}^{d}.
        \end{equation}
    \item Enforcing the touchdown stance leg relative velocity to zero through a forward kinematic constraint:
        \begin{equation}
             \dot{\bm{p}}(^{\text{stance}}\bm{\Gamma}\bm{t}(t_{f}) ,^{\text{stance}}\bm{\Gamma}\dot{\bm{t}}(t_{f})) - \dot{\bm{p}}_{G}(t_{f}) = \bm{0}.
        \end{equation}
    \item Enforcing the liftoff swing leg velocity to zero:
        \begin{equation}
             \dot{\bm{p}}(^{\text{swing}}\bm{\Gamma}\bm{t}(0), ^{\text{swing}}\bm{\Gamma}\dot{\bm{t}}(0)) = \bm{0}.
        \end{equation}
    \item Minimum or desired clearance between the next stance leg and the ground at the beginning of the flight phase: 
        \begin{equation}
            \bm{p}(^{\text{stance}}\bm{\Gamma}\bm{t}(0)) = h_{\text{stance}}.
        \end{equation}
    \item Minimum or desired clearance between the next swing leg and the ground at the end of the flight phase:
        \begin{equation}
            \bm{p}(^{\text{swing}}\bm{\Gamma}\bm{t}(t_{f})) = h_{\text{swing}}.
        \end{equation}
\end{enumerate}
\begin{figure}[b!]
\centerline{\includegraphics[width=\columnwidth]{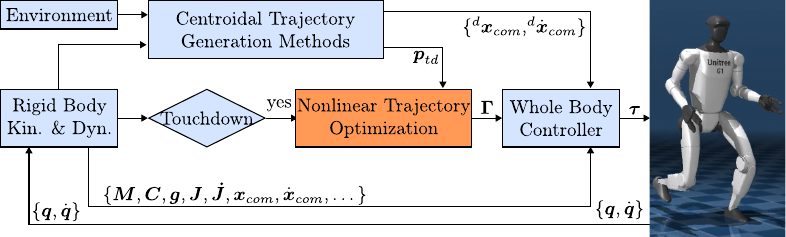}}
\caption{A generic control system diagram of humanoid running or jumping through centroidal trajectory generation methods. This study fits into the orange block which is triggered at the touchdown moment and determines the limb swing trajectories for the flight phase.}
\label{systemDiagram}
\end{figure}
\section{Simulation Results}
The simulation section includes verification of the proposed method through two different humanoid robots. We present the optimization results and then show the optimized trajectories' behavior through a running algorithm. As this study composes a subpart of humanoid running or jumping (see Fig.~\ref{systemDiagram}), we inherit a SLIP-based trajectory generation method from \cite{sovuklukRun}. The running algorithm is combined with the flight phase limb swing trajectories generated by the proposed optimization problem. The overview of the nonlinear optimization solver algorithm that we implemented in C++ can be found in Appendix.
\subsection{Results on Kangaroo}
\begin{figure}[t!]
\centerline{\includegraphics[width=\columnwidth]{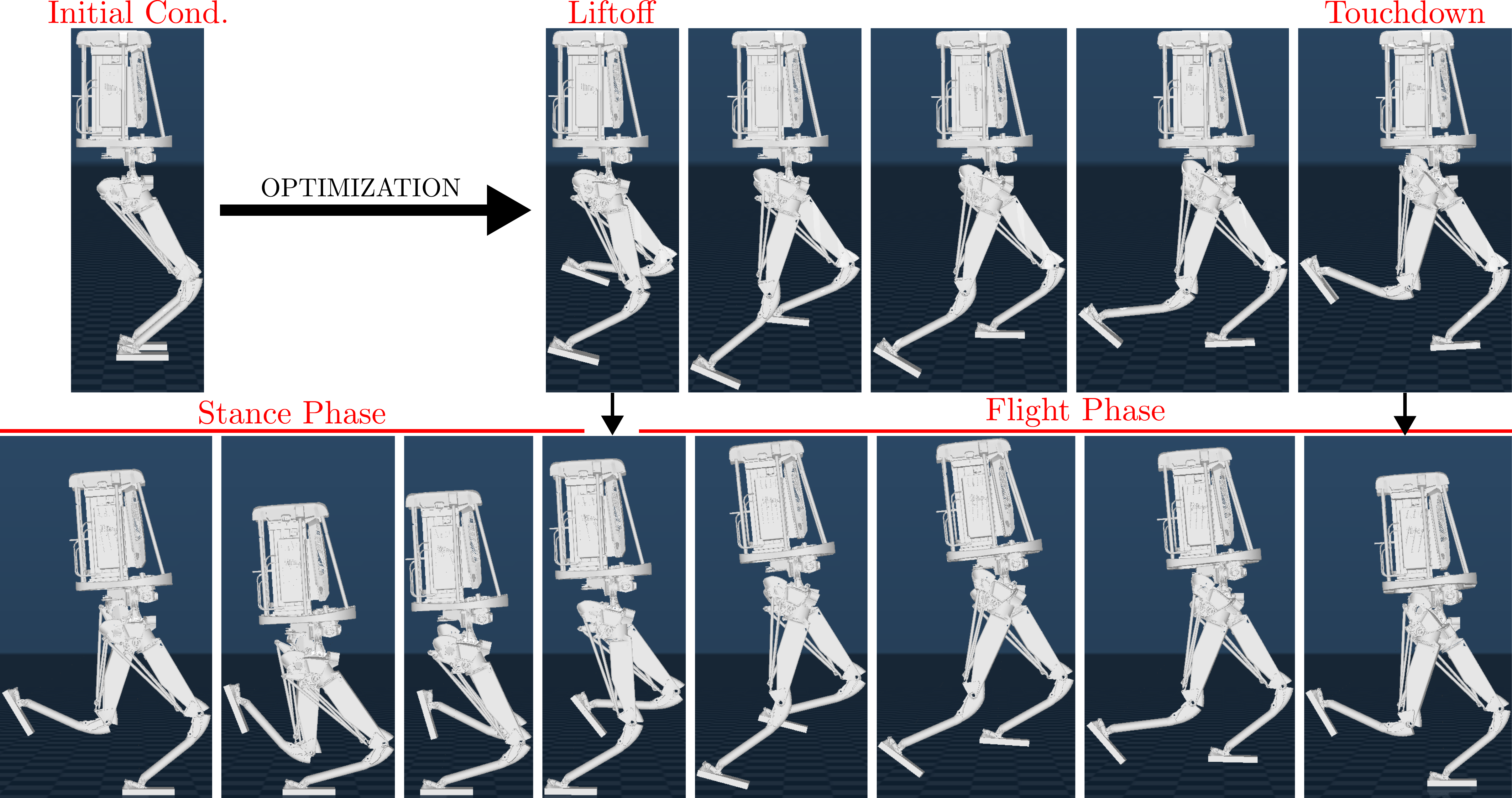}}
\caption{Top row: optimization playback of the flight phase. Bottom row: simulation verification of the optimized trajectories on a running algorithm.}
\label{framesKan}
\end{figure}

We first present the optimization result on the Kangaroo bipedal robot (see Fig.~\ref{framesKan}). Swing leg trajectories of this robot are particularly important as it cannot take advantage of arm swinging for regulation purposes. A running trajectory with $1m/s$ forward velocity, $0.24s$ stance time, and $0.31$ seconds flight time ($\approx$ 12cm of vertical jumping) is obtained from \cite{sovuklukRun} and the optimization problem is configured with the flight time, CoM liftoff velocity, and the relative stance foot locations. It is worth noticing the length of the flight phase of the generated trajectory as it makes postural control harder and causes longer error accumulation in case of imprecision. A 3rd-degree polynomial is employed for each inertially significant joint trajectory. The optimization problem includes 24 optimization parameters along with the 14 nonlinear constraints and samples of 11 points in the dynamics. The optimization problem takes $1.54 ms$ to solve on a daily use desktop computer with AMD Ryzen 7 5800X CPU.

The snapshots of the optimized limb trajectories are shown in the top row of Fig.~\ref{framesKan}. The optimization result suggests that if the robotic system liftoffs with the optimized configuration and velocities and follows the given flight phase swinging behavior, the next touchdown will happen with minimal torso orientation. The verification of the optimized trajectories on the running simulation is shown in the bottom row of Fig.~\ref{framesKan}. The optimization problem is triggered at the moment of touchdown. During the stance phase, the whole-body controller is commanded to liftoff with the optimized configuration and velocities. During the flight phase, the optimized trajectories are followed. The only difference between the optimization playback and the verification simulation is the ankle configuration. Due to its negligible inertial effects, the optimization problem assumes the ankle is always at zero configuration. On the other hand, they are adequately actuated in the simulation. The simulation shows almost identical results with the optimization playback. The optimization playback and running simulation animations are presented in the supplemental video.

\begin{figure}[t!]
\centerline{\includegraphics[width=\columnwidth]{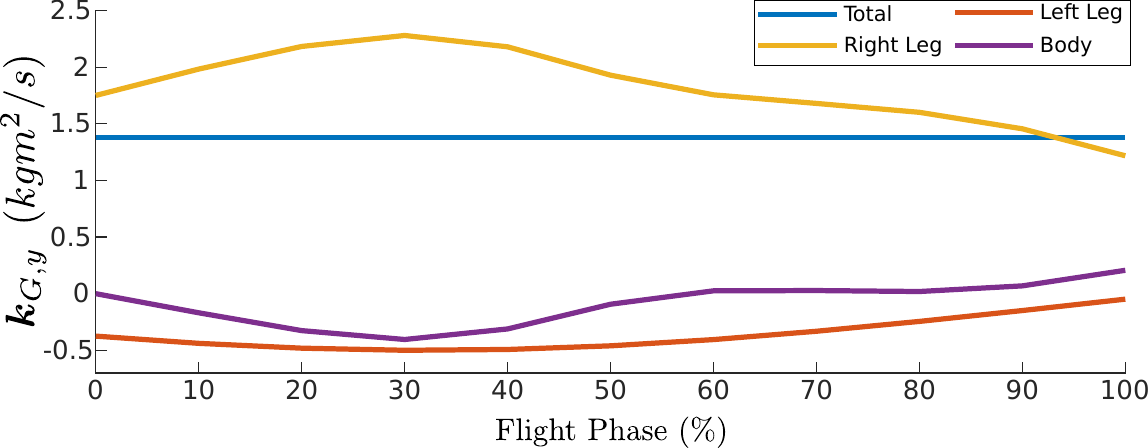}}
\caption{The total centroidal angular momentum in the sagittal plane along with the contribution of each limb. The figure shows that the limb swing trajectories keep the angular momentum portion of the body around zero.}
\label{momentum}
\end{figure}

The first thing to notice on the optimized trajectory is the swing-back behavior of the liftoff leg. This behavior is performed to balance the centroidal angular momentum in the sagittal plane so that the torso movement remains minimal. In order to reason the optimization results, we plot the centroidal angular momentum portions of each limb in the sagittal plane in Fig.~\ref{momentum}. The flight phase angular momentum is also an implicit product of the optimization problem and is constant due to the conservation. As the left leg moves forward, it generates a negative momentum on the CoM frame. The figure shows that the swing-back behavior of the right leg generates a positive momentum to cancel the effect of the other leg's swing. As the left leg reaches its desired position, its velocity and momentum contribution fade away. As the optimization objective is to minimize the body orientation for the next touchdown, the right leg covers most of the angular momentum and keeps the remaining momentum for the torso around zero. In the case of a tilted liftoff condition, the limbs do not cancel the whole momentum and let the body rotate back to the minimal body orientation. Such behavior can also be used for disturbance rejection purposes.
\subsection{Results on Unitree G1}
In order to show the generalizability and inclusiveness of the proposed method, we also optimize for and simulate Unitree's G1 humanoid robot. A running trajectory with 1m/s forward velocity, 0.21s stance time, and 0.26 seconds flight time is generated, and the optimization problem is configured with the flight time, CoM liftoff velocity, and the relative stance foot locations. A 3rd-degree polynomial is employed for each inertially significant leg joint and shoulder pitch joint. The optimization problem includes 32 (24+8) optimization parameters along with the 14 nonlinear constraints and samples of 11 points in the dynamics and takes $1.92 ms$ to solve. The optimization playback and running simulation animations are presented in the supplemental video.

\begin{figure}[b!]
\centerline{\includegraphics[width=\columnwidth]{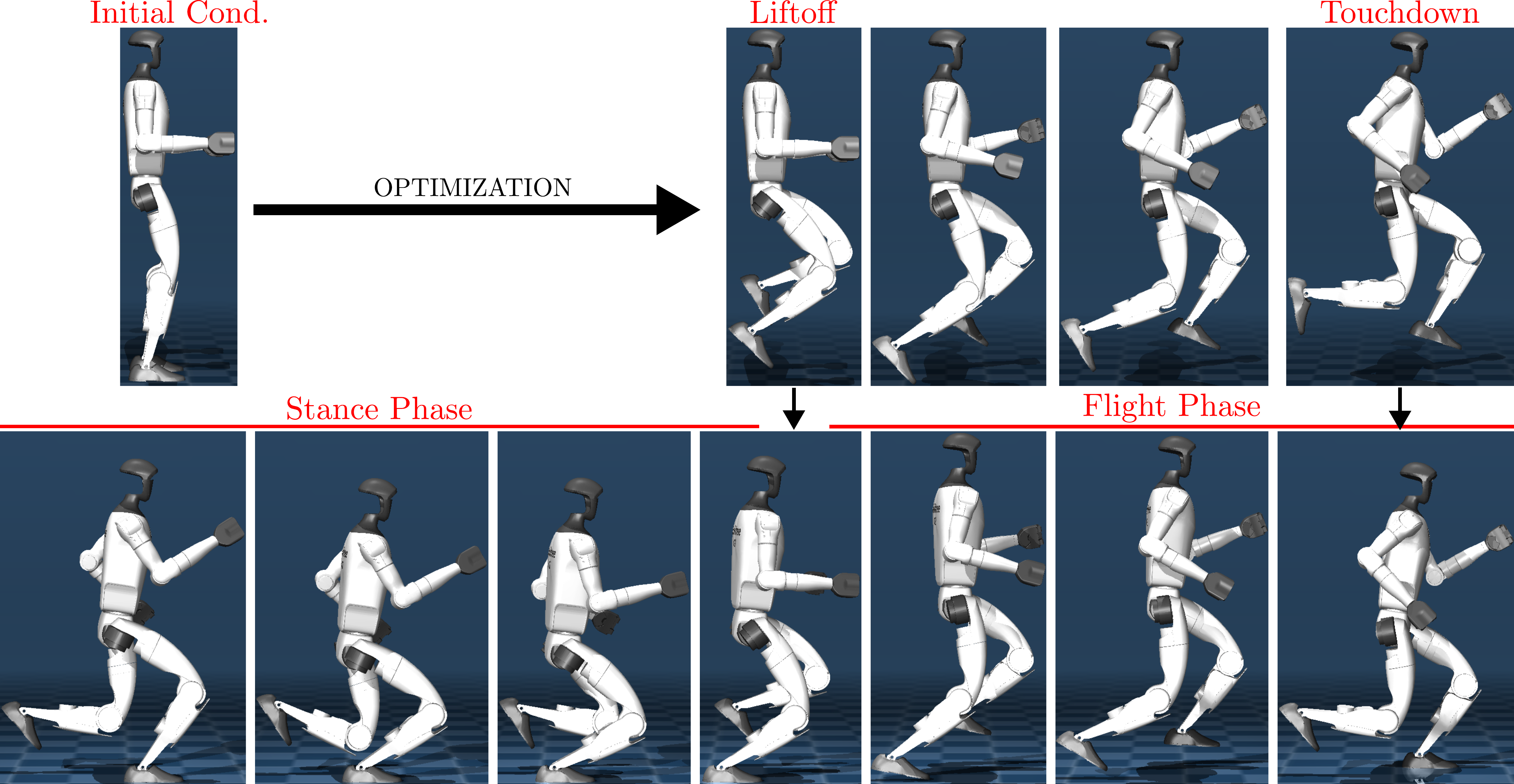}}
\caption{Top row: optimization playback of the flight phase. Bottom row: simulation verification of the optimized trajectories on a running algorithm.}
\label{framesG1}
\end{figure}

A similar swing-back behavior on the right leg can be observed in Fig.~\ref{framesG1}. Differently, the additional arm swing is apparent in the movement. As the right leg covers most of the saggital centroidal angular momentum, the arm movements balance the body rotation in the vertical axis (transverse plane). Even though the optimization is initiated with the zero polynomial configuration, it still manages to converge.
\section{Conclusion}
This study addresses the problem of determining limb swing trajectories for the flight phases of humanoid running. We propose a nonlinear optimization problem to find a set of proper limb swing trajectories that place the stance leg at the desired location and keep the body upright for the next stance phase. We first explore some properties of the angular momentum portion of the centroidal momentum matrix. Then, taking advantage of these properties, we construct a cost function with some constraints for foot placement points. We show that the size of the optimization problem can be significantly reduced through some mild simplifications. We implement the optimized trajectories on a running algorithm from the literature for verification and perform simulations on two different robots. The simulation results verify that once the robotic system lifts off with the optimized configuration and velocities and tracks the optimized flight phase limb trajectories, the robotic system lands with a proper and minimal body orientation. We also share the computational performance results and verify that the proposed optimization problem can be solved in real-time. Lastly, we provide an overview algorithm for the solver that we implement to solve the optimization problem.
\section{Appendix: Solver Algorithm}
We implement our solver, for the given nonlinear optimization problem. For the completeness of the paper, an overview of the solver with the equality constraints is provided in Algorithm~\ref{solver}. The detailed algorithms can be found in \cite{sovukluk2025NFO}. The tool is available at
\begin{equation*} \label{github} \tag{$\star$}
    \text{https://github.com/ssovukluk/ENFORCpp}
\end{equation*}
\begin{algorithm} [htbp]
\caption{Overview of the nonlinear optimizer with equality constraints.}\label{solver}
\small
\begin{enumerate} [wide, labelwidth=!,itemindent=!,labelindent=0pt, leftmargin=0em]
    \item Define $n_{c}$ to be the number of constraints, $f_{c,i}$ to be the $i^{th}$ constraint function, and $\bm{J}$ to be an empty matrix to collect constraint function gradients one by one.
    \item If $n_{c} \neq 0$, set $i = 1$. Otherwise, proceed to step 8.
    \item Calculate numerical gradient of $f_{c,i}$ w.r.t. the optimization variables.
    \item Store the gradient vector: $\bm{J}.\text{col}(i) = \nabla f_{c,i}$.
    \item If $i > 1$, project the estimated gradient into the Nullspace of the vector space $\bm{J}$ such that iterating with this gradient does not disturb the previous constraints.
    \item Implement a line search algorithm to find the zero crossing point of the equality constraint via iterating through the (projected) gradient.
    \item If $i < n_{c}$, set $i = i + 1$ and go to step 3.
\end{enumerate}
\begin{enumerate} [wide, labelwidth=!,itemindent=!,labelindent=0pt, leftmargin=0em]
    \item [*] At this point, all the equality constraints are solved, and the optimizer is ready to proceed to minimize the cost function.
\end{enumerate}
\begin{enumerate} [wide, labelwidth=!,itemindent=!,labelindent=0pt, leftmargin=0em] \setcounter{enumi}{7}
    \item Calculate the numerical gradient of the cost function $f_{\text{cost}}$.
    \item Project the gradient $\nabla f_{\text{cost}}$ into the Nullspace of the vector space $\bm{J}$ such that iterating with this gradient does not disturb the equality constraints.
    \item Implement a line search algorithm to find the local minimum.
    \item Check all the termination conditions and go to step 2 if none of them is triggered.
\end{enumerate}
\end{algorithm}
\bibliographystyle{IEEEtran}
\bibliography{references}
\end{document}